\documentclass[twocolumn, switch]{article} 

\usepackage{preprint}
\usepackage{algorithm}
\usepackage{algorithmic}
\usepackage{amsmath, amsthm, amssymb, amsfonts}
\usepackage{colortbl}
\usepackage{color,xcolor}
\usepackage{booktabs}
\usepackage{multirow}
\usepackage{tikz}
\usetikzlibrary{calc,patterns}

\definecolor{red}{RGB}{202,0,0}
\definecolor{blue}{RGB}{0,102,204}
\definecolor{green}{RGB}{50, 205, 50}
\definecolor{darkorange}{RGB}{255, 140, 0}
\definecolor{lightorange}{RGB}{255, 200, 100}
\definecolor{blue1}{RGB}{230, 240, 250}
\definecolor{blue2}{RGB}{200, 220, 245}
\definecolor{blue3}{RGB}{170, 200, 240}
\definecolor{blue4}{RGB}{140, 180, 235}
\definecolor{blue5}{RGB}{110, 160, 230}
\definecolor{blue6}{RGB}{80, 140, 225}
\definecolor{blue7}{RGB}{50, 120, 220}
\definecolor{blue8}{RGB}{30, 100, 210}
\definecolor{blue9}{RGB}{20, 80, 200}
\definecolor{blue10}{RGB}{0, 60, 190}

\usepackage[numbers,square]{natbib}
\usepackage[utf8]{inputenc}	
\usepackage[T1]{fontenc}	
\usepackage{xcolor}		
\usepackage[colorlinks = true,
            linkcolor = purple,
            urlcolor  = blue,
            citecolor = cyan,
            anchorcolor = black]{hyperref}	
\usepackage{booktabs} 		
\usepackage{nicefrac}		
\usepackage{microtype}		
\usepackage{lineno}		
\usepackage{float}			

\usepackage{lipsum}		

\usepackage{newfloat}
\DeclareFloatingEnvironment[name={Supplementary Figure}]{suppfigure}
\usepackage{sidecap}
\sidecaptionvpos{figure}{c}

\usepackage{titlesec}
\titlespacing\section{0pt}{12pt plus 3pt minus 3pt}{1pt plus 1pt minus 1pt}
\titlespacing\subsection{0pt}{10pt plus 3pt minus 3pt}{1pt plus 1pt minus 1pt}
\titlespacing\subsubsection{0pt}{8pt plus 3pt minus 3pt}{1pt plus 1pt minus 1pt}

\title{Distill, Diffuse, Segment: Unsupervised 3D Semantic Segmentation for Autonomous Driving Based on Multi-Level Distillation and Graph Diffusion}

\usepackage{eso-pic}
\usepackage{tikz}
\usepackage{xcolor}
\PassOptionsToPackage{colorlinks=true,linkcolor=gray,urlcolor=gray}{hyperref}

\newcommand{\AddMyWatermarks}{%
  \begin{tikzpicture}[remember picture, overlay]
    \node[color=gray!90, scale=1] at ([xshift=0in,yshift=-5in]current page.center) {
    };
  \end{tikzpicture}%
}

\AddToShipoutPictureBG*{\AddMyWatermarks}

\usepackage{titling}
\usepackage{orcidlink}
\usepackage{footmisc}
\newcommand{\Author}[3]{
  \textbf{#1}\textsuperscript{#2},\ \orcidlink{#3} %
}

\author{
  \Author{Yijing Wang}{1}{0009-0006-8715-8020} \and
  \Author{Ruonan Li}{2}{0000-0002-9295-7322}\and
  \Author{Qilin Wang}{1}{0009-0002-7481-2995}\and
  \Author{Rongqiang Zhao}{1}{0000-0003-0131-9542} \and
  \Author{Jie Liu}{1}{0000-0001-6209-6886} 
}

\date{%
  \textsuperscript{1}Faculty of Computing, Harbin Institute of Technology\\
  \textsuperscript{2}Pengcheng Laboratory\\[1em]
  \footnotesize \textbf{Corresponding author:} Rongqiang Zhao\texttt{<zhaorq@hit.edu.cn>}
}

\begin{document}

\twocolumn[ 
  \begin{@twocolumnfalse} 

\maketitle
\thispagestyle{empty}

\begin{abstract}
LiDAR-based semantic segmentation is essential for autonomous-driving perception, yet dense point-wise annotations are costly, and long-tailed outdoor scenes make small safety-critical objects difficult to discover without supervision. Existing unsupervised methods face three key challenges: they struggle to preserve small and sparsely observed objects under substantial scale variation, have difficulty enforcing intra-region consistency and inter-region discrimination during cross-modal transfer, and lack an efficient feature-preserving mechanism for contextual propagation over superpoint graphs. We therefore propose DDS, an unsupervised 3D semantic segmentation framework. First, a coarse-to-fine multi-granularity mask cascade provides complementary 3D region cues for objects across different scales, improving the preservation of small and sparsely observed objects. Second, region-guided multi-level distillation transfers self-supervised visual knowledge through point-level alignment, mask-level prototype alignment, and prototype-level contrastive learning, enhancing intra-region consistency and inter-region discrimination. Third, restart-based graph diffusion efficiently propagates contextual information among superpoints while anchoring the refined representation to the initial distilled features and avoiding explicit graph eigendecomposition. Experiments on real-world driving datasets show that DDS outperforms representative unsupervised baselines, improving oAcc, mAcc, and mIoU by up to 2.9\%, 9.7\%, and 4.1\%, respectively. These results demonstrate the effectiveness and transferability of DDS for unsupervised 3D scene understanding in autonomous-driving scenarios.
\end{abstract}
\vspace{0.35cm}

  \end{@twocolumnfalse} 
]

\section{Introduction}
\label{sec:introduction}

LiDAR-based 3D semantic segmentation is essential for autonomous-driving perception, enabling geometric understanding of roads, vehicles, pedestrians, and surrounding infrastructure~\cite{behley2019semantickitti,caesar2020nuscenes}. However, dense point-wise annotations are expensive, while outdoor scenes exhibit severe class imbalance and long-tailed distributions~\cite{cong2021longtail}. Large structures dominate LiDAR observations, whereas safety-critical objects such as pedestrians, bicyclists, and barriers occupy small regions with sparse points. Unsupervised segmentation is therefore attractive for exploiting large-scale unlabeled driving data.

Existing unsupervised 3D segmentation methods mainly follow three
directions. Geometry-driven methods, such as GrowSP and U3DS$^3$,
construct and progressively refine spatially homogeneous
regions~\cite{zhang2023growsp,liu2024u3ds3}. Cross-modal methods transfer
visual representations into 3D; PointDC projects multi-view image
features, while LogoSP combines 2D-to-3D distillation with
graph-frequency-based grouping~\cite{chen2023pointdc,zhang2025logosp}.
Prototype-based methods, such as P-SLCR, improve iterative pseudo-label
learning through structural prototype modeling~\cite{zhan2026pslcr}.
Nevertheless, three limitations remain particularly evident in complex
driving scenes.
\textbf{(1) Inadequate separation of objects across scales.}
Geometry-dominated or single-scale region construction may absorb small
and sparsely observed objects into surrounding dominant regions, while
overly fine partitions can fragment large structures.
\textbf{(2) Insufficient region-level representation learning.}
Predominantly point-wise cross-modal transfer does not explicitly enforce
consistency among points within the same region or discrimination between
different regions, leading to fragmented and weakly separable
representations.
\textbf{(3) Inefficient and insufficiently feature-preserving graph
refinement.}
Existing graph-frequency processing requires explicit spectral
decomposition and lacks a direct mechanism for propagating contextual
information while anchoring the refined features to the transferred
semantics, which may increase computational cost and weaken the learned
representation.

To address these challenges, we propose \emph{Distill, Diffuse, Segment} (DDS), an unsupervised 3D semantic segmentation framework requiring no manual 2D or 3D semantic annotations. First, category-agnostic masks generated at multiple granularities are integrated through a coarse-to-fine cascade and projected into 3D, providing complementary region cues for large structures and small objects. Second, the projected masks organize self-supervised visual knowledge transfer through point-level alignment, mask-level prototype alignment, and prototype-level contrastive learning. Third, the distilled features are aggregated into superpoints and refined by restart-based graph diffusion, which propagates contextual information while preserving the initial representation. Images are used only during distillation, and the final model performs point-cloud-only inference. Experiments on real-world driving datasets demonstrate consistent improvements over representative unsupervised baselines, particularly in class-balanced performance and sparsely observed categories.

The major contributions of this article are as follows:
\begin{itemize}
\item We introduce a multi-granularity region construction strategy that integrates complementary category-agnostic masks through a coarse-to-fine cascade and projects them into 3D, providing scale-aware cues without semantic annotations or text prompts.

\item We propose a region-guided multi-level cross-modal distillation objective that transfers self-supervised visual knowledge through point-level feature alignment, mask-level prototype alignment, and prototype-level contrastive learning, improving intra-region consistency and inter-region discrimination.

\item We develop a restart-based graph-diffusion module that propagates contextual information among superpoints while preserving the distilled representation and avoiding explicit graph eigendecomposition.

\end{itemize}

The remainder of this paper is organized as follows. Related Work reviews existing studies, Method presents the proposed framework, Experiments reports the evaluation results, and Conclusion summarizes the main findings.

\section{Related Work}
\label{sec:related_work}


\noindent\textbf{Weakly Supervised Semantic Segmentation.}
Weakly supervised methods reduce labeling effort by learning from sparse points, clicks, scribbles, partially annotated scenes, or other inexpensive supervisory signals. Early approaches propagate sparse supervision through geometric smoothness, region mining, self-training, or feature redistribution~\cite{xu2020weakly,wei2020multipath,liu2021onething,zhang2021perturbed,wei2024dense}. SQN expands extremely sparse labels through semantic queries, while Scribble-Supervised LiDAR Segmentation and LESS design annotation-efficient supervision for outdoor LiDAR scenes~\cite{hu2022sqn,unal2022scribble,liu2022less}. Other studies improve limited supervision through multi-prototype learning, adaptive perturbation consistency, cross-modal association, and contextual masked modeling~\cite{su2022multiprototype,wu2022dat,sun2022association,liu2023cpcm}. These methods substantially reduce annotation cost, but they still require manually supplied semantic labels and therefore differ from the unsupervised setting considered in this work.

\noindent\textbf{Unsupervised Semantic Segmentation.}
Unsupervised semantic segmentation aims to discover semantic partitions without manual category annotations. A basic baseline clusters handcrafted or learned point descriptors using K-means~\cite{lloyd1982least}. Early 3D evaluations also adapt image-based clustering frameworks, including IIC and PiCIE, to point or superpoint representations~\cite{ji2019iic,cho2021picie}. Meanwhile, self-supervised representation learning methods such as PointContrast, DINO, STEGO, and Masked Scene Contrast provide transferable visual or geometric features, but do not by themselves resolve the complete point-wise semantic discovery problem~\cite{xie2020pointcontrast,caron2021dino,hamilton2022stego,wu2023maskedscene}. Recent methods increasingly combine learned representations with spatial
grouping and iterative pseudo-label refinement.
GrowSP progressively merges geometrically homogeneous superpoints and
clusters their representations to generate semantic pseudo
labels~\cite{zhang2023growsp}.
PointDC transfers multi-view image features into 3D and performs iterative
super-voxel clustering, establishing a representative cross-modal
unsupervised pipeline~\cite{chen2023pointdc}.
U3DS$^3$ combines geometry-based superpoints, spatial clustering,
iterative pseudo-label learning, and transformation consistency without
requiring image pretraining~\cite{liu2024u3ds3}.
LogoSP incorporates 2D-to-3D knowledge distillation to enrich point
features and subsequently performs local--global grouping by analyzing
superpoint representations in the graph-frequency
domain~\cite{zhang2025logosp}.
P-SLCR maintains consistent and ambiguous prototype libraries and reasons
over their semantic relations to mitigate the influence of noisy pseudo
labels~\cite{zhan2026pslcr}. Despite this progress, existing methods remain limited by scale-insensitive region construction, predominantly point-wise visual
transfer, and graph refinement that lacks an explicit mechanism for
preserving the learned representation during contextual propagation.
\begin{figure*}[t]
    \centering
    \IfFileExists{image/Seg_overview.pdf}{%
        \includegraphics[width=\textwidth]{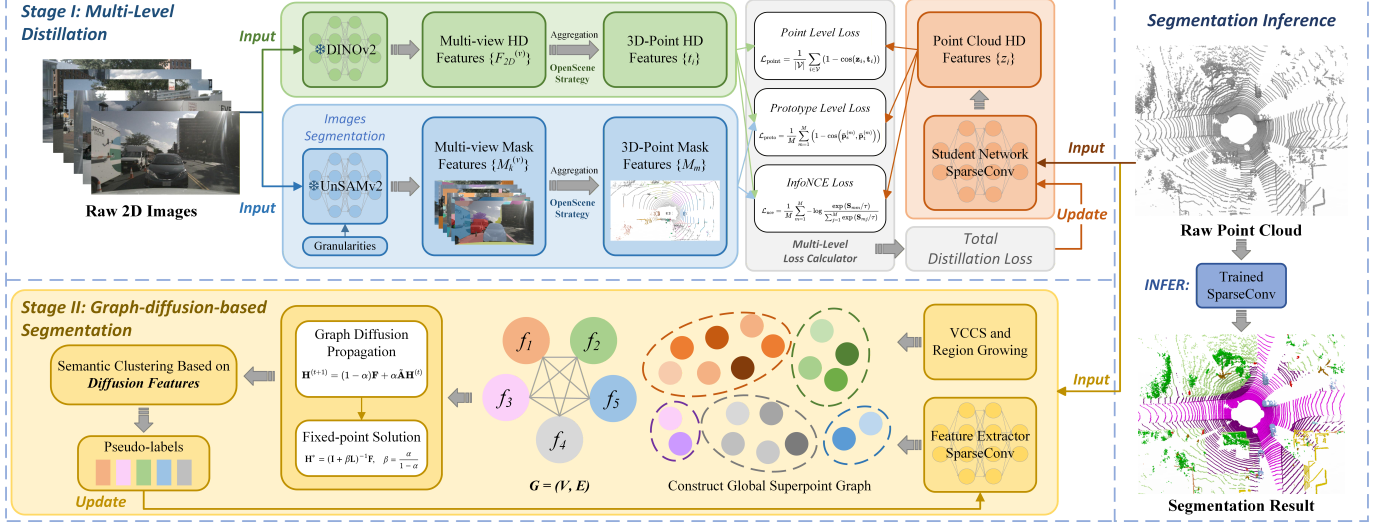}%
    }{%
        \fbox{\parbox[c][4.0cm][c]{0.95\textwidth}{\centering
        Placeholder for the revised DDS pipeline figure.\\
        Upload \texttt{image/Seg\_overview.pdf} to replace this box.}}%
    }
    \caption{Overview of DDS. UnSAMv2 generates complementary masks through a coarse-to-fine granularity cascade, while DINOv2 provides dense self-supervised visual features. The projected masks organize cross-modal distillation at the point, mask-prototype, and inter-prototype levels. The distilled 3D backbone is then incorporated into graph-diffusion-based unsupervised segmentation. Images are required only during training.}
    \label{fig:pipeline}
\end{figure*}

\section{Method}
\label{sec:method}

\subsection{Overview}
\label{subsec:overview}

As illustrated in Fig.~\ref{fig:pipeline}, DDS consists of two training stages. 
The overall framework is unsupervised and does not rely on manual 2D or 3D semantic annotations.

\noindent\textbf{Stage I: Multi-granularity mask construction and multi-level distillation.}
We use frozen UnSAMv2~\cite{yu2025unsamv2} to generate category-agnostic image masks at multiple granularities, while frozen DINOv2~\cite{oquab2023dinov2} provides dense self-supervised visual features. 
The generated masks are projected into 3D to form region groups, which are then combined with the projected DINOv2 features to supervise a sparse 3D backbone. 
The backbone is optimized through a multi-level distillation objective consisting of point-level feature alignment, mask-level prototype alignment, and prototype-level contrastive learning. 
Here, \emph{multi-granularity} refers to the segmentation scales controlled by UnSAMv2, whereas \emph{multi-level} refers to the three levels of knowledge transfer.

\noindent\textbf{Stage II: Graph-diffusion-based unsupervised segmentation.}
The backbone learned in Stage I is used to initialize the segmentation network and extract point features for superpoint construction. 
These features are aggregated into superpoints, organized as a graph, and refined through restart-based graph diffusion~\cite{gasteiger2019predict} to propagate contextual information between related regions. 
The diffused representations are subsequently clustered to generate pseudo labels, which are used to iteratively optimize the segmentation network. 
UnSAMv2 and DINOv2 are required only during Stage I; after training, the final model performs point cloud segmentation using point clouds alone.

\subsection{Multi-Granularity UnSAMv2 Mask Construction}
\label{subsec:unsam_masks}

A single segmentation scale is insufficient for autonomous-driving scenes: coarse masks preserve large structures, whereas fine masks are more likely to isolate small objects. We therefore apply UnSAMv2 at several granularity values and combine its predictions through a coarse-to-fine cascade. The resulting masks are category-agnostic and do not introduce semantic labels or text prompts.

\subsubsection{Coarse-to-Fine Mask Cascade}

Let $\mathcal{I}=\{I^{(v)}\}_{v=1}^{V}$ denote the synchronized camera views. For image $I^{(v)}$, the frozen UnSAMv2 mask generator $\mathcal{U}$ produces a candidate set $\mathcal{R}_{\ell}^{(v)}=\mathcal{U}(I^{(v)};\rho_{\ell})$ at granularity $\rho_{\ell}$. We arrange $\rho_{1}>\rho_{2}>\cdots>\rho_{L}$ from coarse to fine and initialize $\mathcal{K}_{1}^{(v)}=\mathcal{R}_{1}^{(v)}$. At a finer granularity, a candidate is retained only if it does not overlap any mask accepted at coarser granularities:
\begin{equation}
\mathcal{K}_{\ell}^{(v)}=
\left\{
\Omega\in\mathcal{R}_{\ell}^{(v)}
\,\middle|\,
\Omega\cap\Omega'=\varnothing,
\ \forall\Omega'\in\bigcup_{r<\ell}\mathcal{K}_{r}^{(v)}
\right\}.
\label{eq:cascade_suppression}
\end{equation}
The final collection is $\mathcal{K}^{(v)}=\bigcup_{\ell=1}^{L}\mathcal{K}_{\ell}^{(v)}$. This ordering prioritizes structurally complete coarse regions and uses finer-grained masks only to complement spatial regions that remain uncovered, thereby reducing the risk of fragmenting complete objects into local parts. In our implementation, the granularity sequence is $\{0.9,0.7,0.5,0.3\}$.

\subsubsection{Projection and Cross-View Fusion}

Following the calibrated point--pixel association strategy of
OpenScene~\cite{peng2023openscene}, we establish correspondences between
3D points and pixels in all synchronized camera views using the camera
intrinsics and extrinsics.
The same correspondences are used for both UnSAMv2 mask projection and
DINOv2 feature sampling.

Let $\mathcal{P}=\{\mathbf{x}_{i}\}_{i=1}^{N}$ denote the input point
cloud.
For a retained UnSAMv2 mask $\Omega_{k,\ell}^{(v)}$ from view $v$ and
granularity level $\ell$, the corresponding 3D point set is
\begin{equation}
\mathcal{M}_{k,\ell}^{(v)}
=
\left\{
\mathbf{x}_{i}
\,\middle|\,
\delta_{i}^{(v)}=1,\,
\pi^{(v)}(\mathbf{x}_{i})
\in
\Omega_{k,\ell}^{(v)}
\right\},
\label{eq:mask_projection}
\end{equation}
where $\pi^{(v)}(\cdot)$ denotes the calibrated 3D-to-2D projection and
$\delta_i^{(v)}\in\{0,1\}$ indicates whether point $\mathbf{x}_i$ is
visible in view $v$.

We independently project the masks from all camera views and subsequently
fuse them in 3D.
Masks with insufficient point support are removed.
Within the same granularity level, masks whose 3D intersection-over-union
exceeds a predefined threshold are regarded as duplicate observations
and merged.

Because one point may be covered by multiple projected masks, we resolve
the ambiguity using a lexicographic priority.
Masks from finer granularity levels are preferred, while mask confidence
is used to distinguish candidates at the same level.
After cross-view fusion and conflict resolution, we obtain the valid
category-agnostic 3D mask groups
$\mathbb{M}=\{\mathcal{M}_{m}\}_{m=1}^{M}$.
We use $g(i)\in\{-1,1,\ldots,M\}$ to denote the mask assigned to point
$i$, where $g(i)=-1$ indicates that the point is not assigned to any
valid mask.

\begin{figure}[t]
    \centering
    \IfFileExists{image/3D-mask.pdf}{%
        \includegraphics[width=0.95\columnwidth]{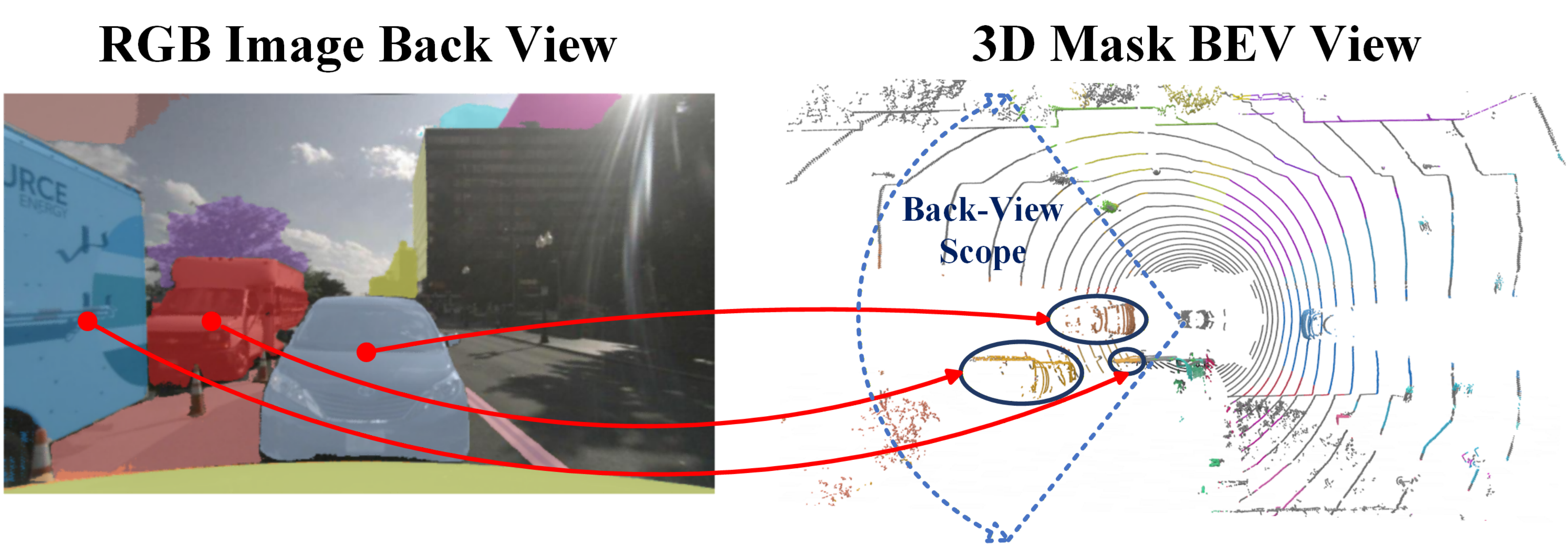}%
    }{%
        \fbox{\parbox[c][3.2cm][c]{0.9\columnwidth}{\centering
        Placeholder for multi-granularity mask projection.\\
        Upload \texttt{image/3D-mask.pdf} to replace this box.}}%
    }
    \caption{Multi-granularity UnSAMv2 masks are projected from synchronized camera views and fused into category-agnostic 3D mask groups.}
    \label{fig:3dmask}
\end{figure}

\subsection{Multi-Level Cross-Modal Distillation}
\label{subsec:distill}

The projected UnSAMv2 masks specify \emph{where} region consistency should be imposed, but they do not provide semantic category labels or dense representations. DINOv2 complements these masks by specifying \emph{what} visual information should be transferred. We combine both cues to supervise the 3D backbone at three representation levels.

\subsubsection{Teacher Features and Mask Prototypes}

Using the same point--pixel correspondences, we follow the multi-view
feature fusion strategy of OpenScene~\cite{peng2023openscene} to transfer
dense DINOv2 features from images to 3D points.
For each view $v$, the frozen DINOv2 encoder extracts a dense feature map
$\mathbf{F}_{\mathrm{2D}}^{(v)}
\in\mathbb{R}^{C\times H\times W}$.
Projecting point $\mathbf{x}_i$ into view $v$ gives the corresponding
sampled feature
$\mathbf{f}_{i}^{(v)}\in\mathbb{R}^{C}$.

We average all valid observations associated with the same point to
construct its multi-view teacher feature:
\begin{equation}
\mathbf{t}_{i}
=
\frac{
\sum_{v=1}^{V}
\delta_{i}^{(v)}
\mathbf{f}_{i}^{(v)}
}{
\sum_{v=1}^{V}
\delta_{i}^{(v)}
+
\varepsilon
},
\label{eq:teacher_feature}
\end{equation}
where $\varepsilon$ is a small constant for numerical stability and
$\mathcal{V}=\{i\mid\sum_{v}\delta_i^{(v)}>0\}$ denotes the set of points
visible in at least one view.
After voxelization, the sparse 3D student backbone produces
$\mathbf{z}_{i}=f_{\theta}(\mathbf{x}_{i})\in\mathbb{R}^{C}$.

For mask $m$, let
$\mathcal{V}_{m}=\{i\in\mathcal{V}\mid g(i)=m\}$
denote its assigned visible points.
We aggregate the student and teacher features within the same projected
mask to construct paired region prototypes:
\begin{equation}
\mathbf{p}^{(m)}_{\mathrm{s}}
=
\frac{1}{|\mathcal{V}_{m}|}
\sum_{i\in\mathcal{V}_{m}}
\mathbf{z}_{i},
\qquad
\mathbf{p}^{(m)}_{\mathrm{t}}
=
\frac{1}{|\mathcal{V}_{m}|}
\sum_{i\in\mathcal{V}_{m}}
\mathbf{t}_{i}.
\label{eq:prototypes}
\end{equation}
Masks containing too few visible points are discarded to reduce noise
caused by sparse observations and imperfect cross-view projection.
The resulting prototypes provide paired region-level representations for
cross-modal alignment, without introducing semantic category labels.

\subsubsection{Distillation Objective}

The distillation objective consists of three complementary terms: point-level feature alignment, mask-level prototype alignment, and prototype-level contrastive learning.
For point-level distillation, each student feature $\mathbf{z}_i$ is aligned with its corresponding DINOv2 teacher feature $\mathbf{t}_i$.
For mask-level distillation, we use the normalized student and teacher prototypes $\bar{\mathbf{p}}^{(m)}_{\mathrm{s}}$ and $\bar{\mathbf{p}}^{(m)}_{\mathrm{t}}$ computed from each valid 3D mask.
To encourage inter-mask separability, we further construct a prototype similarity matrix
$\mathbf{S}=\bar{\mathbf{P}}_{\mathrm{s}}\bar{\mathbf{P}}_{\mathrm{t}}^{\top}$,
where $\bar{\mathbf{P}}_{\mathrm{s}}, \bar{\mathbf{P}}_{\mathrm{t}} \in \mathbb{R}^{M\times C}$ are the stacked normalized student and teacher prototypes.

The three loss terms are defined as
\begin{align}
\mathcal{L}_{\mathrm{point}}
&=
\frac{1}{|\mathcal{V}|}
\sum_{i\in\mathcal{V}}
\left(1-\cos(\mathbf{z}_i,\mathbf{t}_i)\right),
\label{eq:lpoint}\\
\mathcal{L}_{\mathrm{proto}}
&=
\frac{1}{M}
\sum_{m=1}^{M}
\left(
1-\cos\!\left(
\bar{\mathbf{p}}^{(m)}_{\mathrm{s}},
\bar{\mathbf{p}}^{(m)}_{\mathrm{t}}
\right)
\right),
\label{eq:lproto}\\
\mathcal{L}_{\mathrm{nce}}
&=
\frac{1}{M}
\sum_{m=1}^{M}
-\log
\frac{
\exp\left(\mathbf{S}_{mm}/\tau\right)
}{
\sum_{j=1}^{M}\exp\left(\mathbf{S}_{mj}/\tau\right)
},
\label{eq:lnce}
\end{align}
where $\tau$ is the temperature parameter.
In $\mathcal{L}_{\mathrm{nce}}$, the teacher prototype with the same mask index is treated as the positive sample, while the remaining teacher prototypes are treated as negatives.

The final distillation objective is
\begin{equation}
\mathcal{L}_{\mathrm{distill}}
=
\lambda_{\mathrm{point}}\mathcal{L}_{\mathrm{point}}
+
\lambda_{\mathrm{proto}}\mathcal{L}_{\mathrm{proto}}
+
\lambda_{\mathrm{nce}}\mathcal{L}_{\mathrm{nce}}.
\label{eq:ldistill}
\end{equation}
By minimizing Eq.~\eqref{eq:ldistill}, the 3D backbone learns point-level semantic alignment, region-level consistency, and inter-mask discrimination.
After training, the distilled backbone can extract semantically meaningful point features using only the point cloud at inference time.
\subsection{Graph Diffusion for Unsupervised Segmentation}
\label{subsec:diffusion}

We initialize the segmentation backbone with the distilled checkpoint
instead of random weights.
The learned point features are aggregated into superpoints, propagated
over a relational graph, and clustered to construct pseudo labels for
iterative self-training.

\subsubsection{Superpoint Graph Construction}

We over-segment each point cloud into
$\mathcal{S}=\{S_n\}_{n=1}^{N_s}$
using VCCS~\cite{papon2013vccs}, followed by region
growing~\cite{adams1994seeded}.
The feature of each superpoint is obtained by average pooling the
features of its constituent points.
We then construct a feature-affinity graph using an RBF kernel and
symmetrically normalize its adjacency matrix:
\begin{align}
\mathbf{F}
&=
\left[
\mathbf{h}_{1},
\ldots,
\mathbf{h}_{N_s}
\right]^{\top},
\qquad
\mathbf{h}_{n}
=
\frac{1}{|S_n|}
\sum_{i\in S_n}\mathbf{z}_{i},
\label{eq:superpoint_feature}
\\
A_{ij}
&=
\exp\!\left(
-\gamma
\left\|
\mathbf{h}_{i}-\mathbf{h}_{j}
\right\|_{2}^{2}
\right),
\label{eq:affinity}
\\
\widetilde{\mathbf{A}}
&=
\mathbf{D}^{-\frac{1}{2}}
\mathbf{A}
\mathbf{D}^{-\frac{1}{2}},
\qquad
D_{ii}
=
\sum_{j=1}^{N_s} A_{ij}.
\label{eq:normalized_adj}
\end{align}
Here, $\mathbf{F}\in\mathbb{R}^{N_s\times C}$ stacks all superpoint
features, and $\gamma=1/C$ in our implementation.
The corresponding normalized graph Laplacian is
$\mathbf{L}=\mathbf{I}-\widetilde{\mathbf{A}}$.

\subsubsection{Diffusion and Pseudo-Label Generation}

Starting from $\mathbf{H}^{(0)}=\mathbf{F}$, graph diffusion propagates
contextual information among related superpoints while preserving the
initial representation:
\begin{align}
\mathbf{H}^{(t+1)}
&=
(1-\alpha)\mathbf{F}
+
\alpha\widetilde{\mathbf{A}}\mathbf{H}^{(t)},
\label{eq:diffusion_iter}
\\
\mathbf{H}^{*}
&=
\left(
\mathbf{I}+\beta\mathbf{L}
\right)^{-1}
\mathbf{F},
\qquad
\beta
=
\frac{\alpha}{1-\alpha}.
\label{eq:diffusion_closed}
\end{align}
Here, $\alpha\in(0,1)$ is the diffusion coefficient.
The first term in Eq.~\eqref{eq:diffusion_iter} anchors each update to
the original superpoint features, while the second propagates information
over the graph.
Equation~\eqref{eq:diffusion_closed} gives the fixed-point solution.
In practice, we avoid explicit matrix inversion and use
$\mathbf{H}^{(T)}$ after a fixed number of diffusion iterations as its
efficient approximation. We retain the informative channels of $\mathbf{H}^{(T)}$ according to
their cumulative variance and apply KMeans to obtain coarse groups.
The grouped features are then projected by PCA and clustered again to
form semantic primitives.

For primitive $p$, let $\mathcal{C}_p$ denote its associated superpoint
set.
Its feature center is computed, and each superpoint is reassigned
according to cosine similarity with the normalized primitive centers:
\begin{align}
\mathbf{c}_{p}
&=
\frac{1}{|\mathcal{C}_{p}|}
\sum_{n\in\mathcal{C}_{p}}
\mathbf{h}_{n}^{(T)},
\label{eq:primitive_center}
\\
\widehat{y}_{n}
&=
\arg\max_{p}
\left\langle
\frac{\mathbf{h}_{n}^{(T)}}
{\left\|\mathbf{h}_{n}^{(T)}\right\|_{2}},
\frac{\mathbf{c}_{p}}
{\left\|\mathbf{c}_{p}\right\|_{2}}
\right\rangle.
\label{eq:primitive_assign}
\end{align}
Here, $\mathbf{h}_{n}^{(T)}$ is the $n$-th row of
$\mathbf{H}^{(T)}$.

The superpoint assignments are mapped back to the original points by
$\widehat{y}_{i}=\widehat{y}_{s(i)}$, where $s(i)$ denotes the
superpoint containing $\mathbf{x}_i$.
These assignments serve as pseudo labels for iterative self-training.
The final network predicts point-wise clusters from point clouds alone,
without external semantic category labels.

\begin{figure*}[t]
    \centering
    \includegraphics[width=\textwidth]{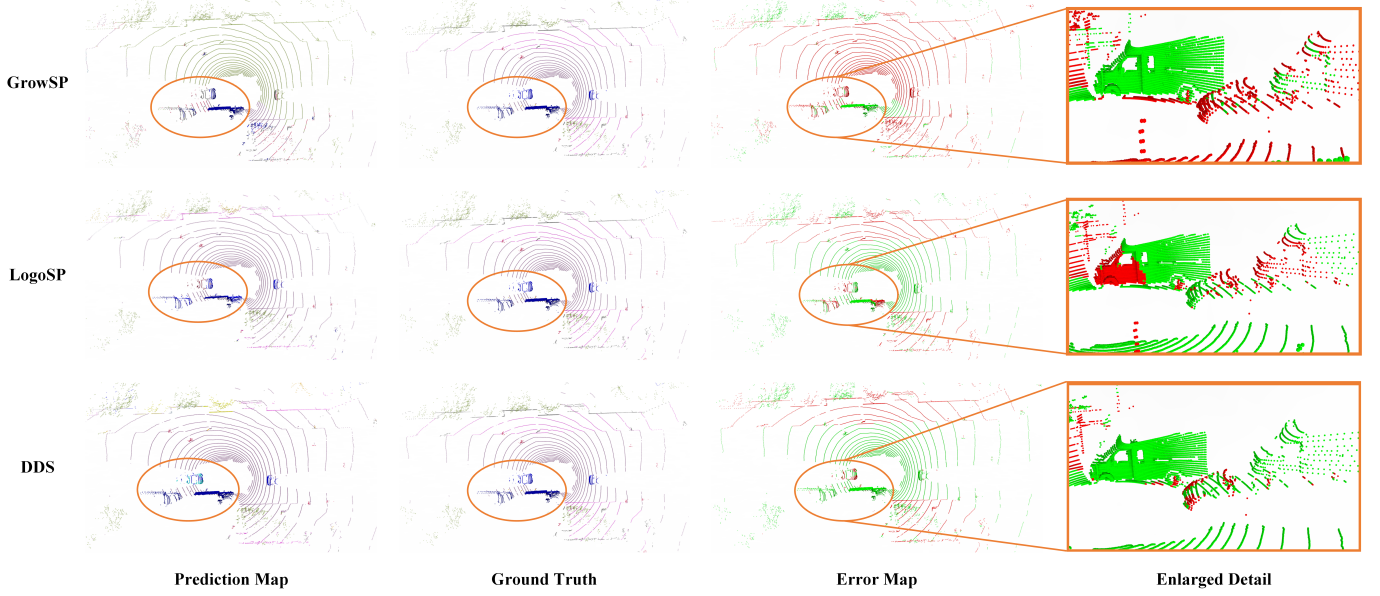}
    \caption{
    Visualization comparison of GrowSP, LogoSP, and DDS on a representative
    nuScenes frame.
    From left to right, each row shows the prediction map, ground truth,
    error map, and enlarged detail.
    Green and red points in the error map denote correct and incorrect
    predictions, respectively.
    }
    \label{fig:compare_nuscenes}
\end{figure*}
\section{Experiments}
\label{sec:experiments}

\subsection{Implementation}

\subsubsection{Datasets and Evaluation Metrics}
We conduct experiments on two large-scale LiDAR semantic segmentation benchmarks collected in real-world driving environments: \textbf{nuScenes}~\cite{caesar2020nuscenes} and \textbf{SemanticKITTI}~\cite{behley2019semantickitti}.
nuScenes contains 1,000 driving scenes collected with a 32-beam LiDAR, and we evaluate 16 semantic classes.
SemanticKITTI contains more than 43,000 scans collected with a 64-beam LiDAR and defines 19 classes for semantic segmentation evaluation.
We report overall accuracy (oAcc), mean class accuracy (mAcc), and mean Intersection over Union (mIoU).

\subsubsection{Baselines}
We compare DDS with representative unsupervised semantic segmentation methods.
The comparison includes generic clustering baselines, namely K-means~\cite{lloyd1982least}, IIC~\cite{ji2019iic}, and PiCIE~\cite{cho2021picie}; point-cloud-oriented methods, including GrowSP~\cite{zhang2023growsp}, U3DS$^3$~\cite{liu2024u3ds3}, and P-SLCR~\cite{zhan2026pslcr}; and cross-modal methods, including PointDC~\cite{chen2023pointdc}, PointDC-DINOv2~\cite{zhang2025logosp}, and LogoSP~\cite{zhang2025logosp}.
PointDC-DINOv2 denotes the DINOv2-based PointDC variant reported by LogoSP.

\subsubsection{Experimental Settings}

Our framework is implemented in PyTorch with MinkowskiEngine~\cite{choy20194d}
and trained on a single NVIDIA RTX 3090 GPU with 24 GB memory.
Additional implementation details are provided in the Appendix~E.
For SemanticKITTI, LogoSP and DDS follow the same cross-dataset protocol:
both are initialized with their respective 3D backbones distilled on
nuScenes, without using SemanticKITTI images.
LogoSP subsequently applies GFT-based grouping, whereas DDS uses graph
diffusion followed by clustering and iterative self-training.
All subsequent stages use only SemanticKITTI point clouds, evaluating the
cross-dataset transferability of the distilled representations across
different LiDAR sensors.

\begin{table}[t]
    \centering
    {\small
    \setlength{\tabcolsep}{4.5pt}
    \begin{tabular}{l|ccc}
        \toprule
        Method
        & oAcc (\%)$\uparrow$
        & mAcc (\%)$\uparrow$
        & mIoU (\%)$\uparrow$ \\
        \midrule
        K-means~\citeyearpar{lloyd1982least}
        & 10.9 & 13.3 & 3.4 \\
        IIC~\citeyearpar{ji2019iic}
        & 22.7 & 11.5 & 4.6 \\
        PiCIE~\citeyearpar{cho2021picie}
        & 32.8 & 21.0 & 8.0 \\
        GrowSP~\citeyearpar{zhang2023growsp}
        & 39.2 & 17.5 & 10.2 \\
        PointDC~\citeyearpar{chen2023pointdc}
        & 56.8 & 29.4 & 17.7 \\
        PointDC-DINOv2~\citeyearpar{zhang2025logosp}
        & 51.8 & 28.6 & 18.2 \\
        LogoSP~\citeyearpar{zhang2025logosp}
        & 54.8 & 29.2 & 20.1 \\
        \textbf{DDS (Ours)}
        & \textbf{59.6}
        & \textbf{31.5}
        & \textbf{21.5} \\
        \bottomrule
    \end{tabular}
    }
    \caption{Comparison with unsupervised semantic segmentation methods
    on nuScenes. The best result in each column is shown in bold.}
    \label{tab:sota_comparison_nuscenes}
\end{table}
\begin{table}[t]
    \centering
    {\small
    \setlength{\tabcolsep}{4.5pt}
    \begin{tabular}{l|ccc}
        \toprule
        Method
        & oAcc (\%)$\uparrow$
        & mAcc (\%)$\uparrow$
        & mIoU (\%)$\uparrow$ \\
        \midrule
        K-means~\citeyearpar{lloyd1982least}
        & 9.1 & 6.8 & 2.3 \\
        IIC~\citeyearpar{ji2019iic}
        & 28.0 & 5.6 & 2.1 \\
        PiCIE~\citeyearpar{cho2021picie}
        & 29.4 & 6.7 & 2.8 \\
        GrowSP~\citeyearpar{zhang2023growsp}
        & 40.9 & 19.1 & 12.5 \\
        U3DS$^3$~\citeyearpar{liu2024u3ds3}
        & 34.8 & 23.1 & 14.2 \\
        LogoSP~\citeyearpar{zhang2025logosp}
        & 49.2 & 23.1 & 15.5 \\
        P-SLCR~\citeyearpar{zhan2026pslcr}
        & 55.9 & 21.1 & 15.3 \\
        \textbf{DDS (Ours)}
        & \textbf{58.8}
        & \textbf{32.8}
        & \textbf{19.6} \\
        \bottomrule
    \end{tabular}
    }
    \caption{Comparison with unsupervised semantic segmentation methods
    on SemanticKITTI. The best result in each column is shown in bold.}
    \label{tab:sota_comparison_semantic_kitti}
\end{table}
\begin{figure}[t]
    \centering
    \includegraphics[width=\columnwidth]{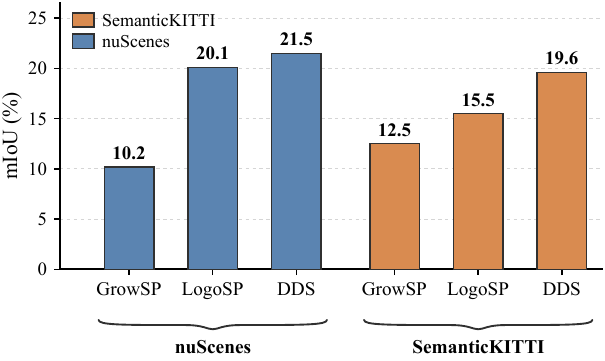}
    \caption{Bar chart of mIoU (\%) among GrowSP, LogoSP, and DDS on nuScenes and SemanticKITTI. DDS achieves the best performance on both datasets.}
    \label{fig:miou_comparison}
\end{figure}
\begin{table}[t]
    \centering
    {\small
    \setlength{\tabcolsep}{2.5pt}
    \begin{tabular}{l|l|ccc}
        \toprule
        Init.
        & Process.
        & oAcc (\%)$\uparrow$
        & mAcc (\%)$\uparrow$
        & mIoU (\%)$\uparrow$ \\
        \midrule
        Rand & PCA & 48.1 & 12.5 & 8.1 \\
        Rand & GFT & 46.6 & 12.6 & 8.4 \\
        Rand & Diffusion & 48.6 & 13.1 & 8.7 \\
        \midrule
        P-distill & PCA & 53.2 & 26.6 & 16.5 \\
        P-distill & GFT & 52.3 & 27.1 & 19.2 \\
        P-distill & Diffusion & 56.9 & 29.5 & 19.7 \\
        \midrule
        M-distill & PCA & 58.7 & 30.5 & 20.6 \\
        M-distill & GFT & 53.0 & 25.7 & 16.3 \\
        \textbf{M-distill} & \textbf{Diffusion}
        & \textbf{59.6} & \textbf{31.5} & \textbf{21.5} \\
        \bottomrule
    \end{tabular}
    }
    \caption{Ablation of distillation initialization and superpoint feature
    processing on nuScenes. The best result in each column is shown in bold. Rand, P-distill, and M-distill denote random initialization,
        point-level distillation, and the proposed multi-level distillation,
        respectively. PCA directly reduces and clusters the superpoint
        features without graph modeling. GFT performs graph-frequency-based
        local--global modeling following LogoSP~\cite{zhang2025logosp}.
        Diffusion denotes our restart-based graph propagation over the
        normalized superpoint affinity graph.}
    \label{tab:ablation_components}
\end{table}

\subsection{Comparison with Unsupervised Methods}

\begin{figure*}[t]
    \centering
    \includegraphics[width=\textwidth]{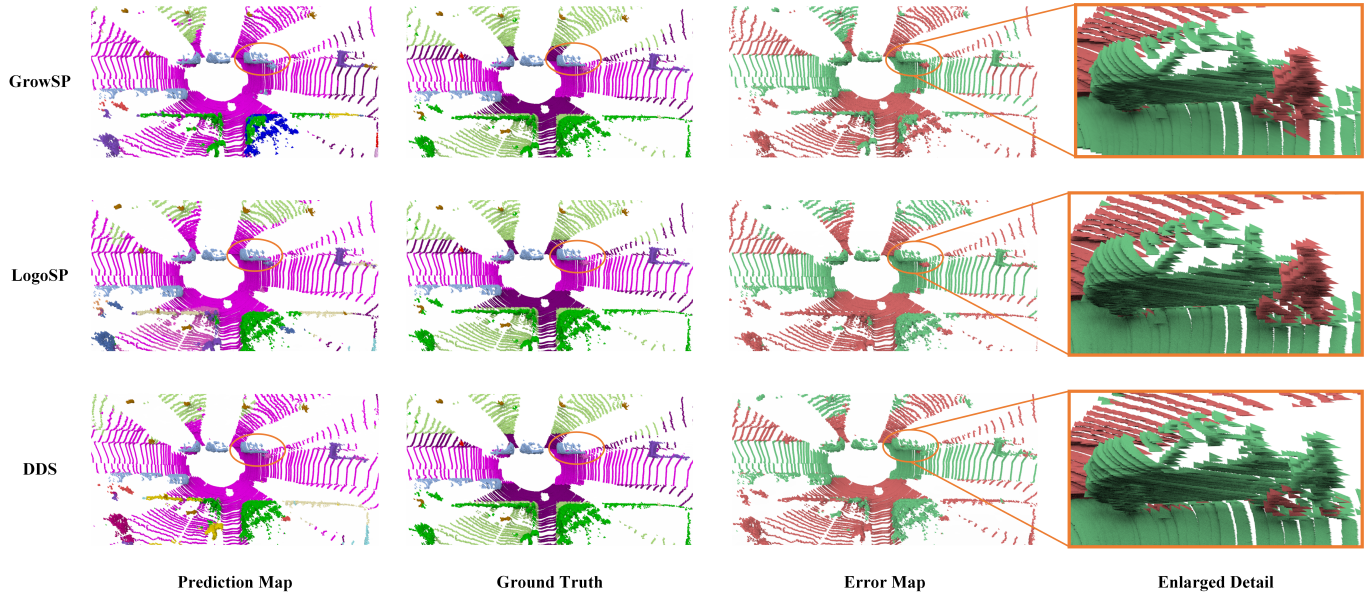}
    \caption{
    Visualization comparison of GrowSP, LogoSP, and DDS on a representative
    SemanticKITTI frame.
    From left to right, each row shows the prediction map, ground truth,
    error map, and enlarged detail.
    Green and red points in the error map denote correct and incorrect
    predictions, respectively.
    }
    \label{fig:compare_semantickitti}
\end{figure*}
\begin{table}[t]
    \centering
    {\small
    \setlength{\tabcolsep}{3pt}
    \begin{tabular}{c|c|ccc}
        \toprule
        $\lambda_{\mathrm{proto}}$
        & $\lambda_{\mathrm{nce}}$
        & oAcc (\%)$\uparrow$
        & mAcc (\%)$\uparrow$
        & mIoU (\%)$\uparrow$ \\
        \midrule
        0 & 0   & 56.9 & 29.5 & 19.7 \\
        1 & 0   & 59.1 & 30.3 & 20.5 \\
        \textbf{1} & \textbf{0.2}
        & \textbf{59.6} & \textbf{31.5} & \textbf{21.5} \\
        1 & 0.4 & 57.6 & 31.1 & 20.5 \\
        1 & 0.6 & 56.6 & 29.5 & 18.1 \\
        \bottomrule
    \end{tabular}
    }
    \caption{Loss-weight ablation for multi-level distillation on nuScenes. We fix $\lambda_{\mathrm{point}}=1$. The best result in each column is shown in bold.}
    \label{tab:distill_lambda_ablation}
\end{table}
\subsubsection{Quantitative Comparison}

Tables~\ref{tab:sota_comparison_nuscenes} and
\ref{tab:sota_comparison_semantic_kitti} report the quantitative
comparisons with representative unsupervised semantic segmentation
methods.

On nuScenes, DDS achieves 59.6\% oAcc, 31.5\% mAcc, and 21.5\% mIoU.
Compared with the strongest baseline for each metric, DDS improves oAcc
by 2.8 percentage points over PointDC, mAcc by 2.1 points over PointDC,
and mIoU by 1.4 points over LogoSP.
These gains demonstrate that DDS improves both overall scene organization
and class-balanced segmentation quality.

On SemanticKITTI, DDS obtains 58.8\% oAcc, 32.8\% mAcc, and 19.6\% mIoU.
It exceeds the strongest competing results by 2.9 points in oAcc,
9.7 points in mAcc, and 4.1 points in mIoU.
Notably, these results are obtained without repeating the visual
distillation stage on SemanticKITTI, indicating that the visually
distilled 3D representation transfers effectively across different
LiDAR datasets.
The substantial improvement in mAcc further suggests that DDS produces
more balanced predictions across semantic categories, rather than mainly
benefiting dominant classes. As shown in Fig.~\ref{fig:miou_comparison}, DDS consistently outperforms the representative unsupervised baselines on both nuScenes and SemanticKITTI, demonstrating its strong generalizability across diverse autonomous-driving scenarios.

The class-wise results further reveal particularly large improvements
for small or sparsely observed categories.
On nuScenes, DDS improves the pedestrian IoU from the second-best
33.6\% to 40.0\%, a gain of 6.4 percentage points.
On SemanticKITTI, the bicyclist IoU increases from 1.1\% to 7.9\%,
while the fence IoU increases from 3.0\% to 11.0\%, corresponding to
gains of 6.8 and 8.0 percentage points, respectively.
These improvements support the effectiveness of DDS in preserving
the structure of small and locally distributed objects. Detailed class-wise IoU results for both datasets are provided in the
Appendix~D.

\subsubsection{Visualization Comparison}

Figures~\ref{fig:compare_nuscenes} and
\ref{fig:compare_semantickitti} provide visualization comparisons among
GrowSP, LogoSP, and DDS on representative nuScenes and SemanticKITTI
frames. These visualizations show that DDS better preserves semantic consistency in challenging local regions, such as closely spaced vehicles and vehicle-cyclist areas, and reduces fragmented or noisy predictions compared with GrowSP and LogoSP.

\subsection{Ablation Studies}

We conduct ablation studies on nuScenes to examine the effects of the
distillation strategy, superpoint feature processing, and loss weights.
Additional ablations on the coarse-to-fine mask cascade are provided in
Appendix~A.

\subsubsection{Component Ablation}

Table~\ref{tab:ablation_components} evaluates the interaction between
feature initialization and superpoint processing.

With Diffusion fixed, point-level distillation improves the Rand setting
from 48.6\%, 13.1\%, and 8.7\% to 56.9\%, 29.5\%, and 19.7\% in oAcc,
mAcc, and mIoU, respectively.
Replacing P-distill with M-distill further improves the three metrics by
2.7, 2.0, and 1.8 percentage points, confirming the additional benefits
of mask-level prototype alignment and prototype-level contrastive
learning. Graph diffusion also provides consistent gains when coupled with the
proposed multi-level distillation.
Under M-distill, Diffusion outperforms PCA by 0.9 points in oAcc,
1.0 point in mAcc, and 0.9 points in mIoU.
It also exceeds GFT by 6.6, 5.8, and 5.2 points, respectively.
These results indicate that graph diffusion more effectively exploits
the distilled representation.

\subsubsection{Loss-Weight Ablation}

Table~\ref{tab:distill_lambda_ablation} studies the weights of the prototype alignment and prototype-level contrastive terms while fixing $\lambda_{\mathrm{point}}=1$.
Using only point-level alignment yields 56.9\% oAcc, 29.5\% mAcc, and 19.7\% mIoU.
Adding prototype alignment increases the metrics to 59.1\%, 30.3\%, and 20.5\%, demonstrating the benefit of region-level consistency.
Introducing the contrastive term with $\lambda_{\mathrm{nce}}=0.2$ produces the best overall result of 59.6\% oAcc, 31.5\% mAcc, and 21.5\% mIoU.
Larger contrastive weights reduce oAcc and mIoU, with $\lambda_{\mathrm{nce}}=0.6$ decreasing mIoU to 18.1\%.
This trend suggests that a moderate contrastive weight provides the best balance between inter-region discrimination and stable region-level alignment.

\section{Conclusion}
\label{sec:conclusion}

In this work, we proposed DDS, an unsupervised framework for 3D
semantic segmentation in autonomous-driving scenes.
DDS combines multi-granularity UnSAMv2 mask construction with multi-level
cross-modal distillation to transfer point- and region-level visual
knowledge into a sparse 3D backbone. Restart-based graph diffusion is further introduced to refine superpoint representations and generate more coherent pseudo labels efficiently. Experiments on driving datasets demonstrate that DDS consistently outperforms representative unsupervised methods and that the distilled
representation can effectively transfer across different LiDAR domains.
Future work will explore temporal information from consecutive LiDAR
scans and more adaptive region construction to improve segmentation of
sparse and long-tailed objects.

\normalsize
\bibliography{main}


\footnotesize

\end{document}